\documentclass[journal]{IEEEtran}
\ifCLASSINFOpdf
\else
\fi
\usepackage{subfigure}
\usepackage{float}
\usepackage{multirow}
\usepackage{multicol} 
\usepackage{xcolor}
\usepackage{colortbl}  %
\usepackage{threeparttable} 
\usepackage{enumerate}
\usepackage{amsmath}
\usepackage{setspace}  
\usepackage{amsmath}
\usepackage{graphicx} 
\usepackage{textcomp}
\usepackage{array}
\begin{document}
%
\title{Spoofing and Anti-Spoofing with Wax Figure Faces}
\author{Shan Jia,
        Xin Li,~\IEEEmembership{Fellow,~IEEE,}
        Chuanbo Hu,
        and~Zhengquan Xu
\thanks{S. Jia and Z. Xu are with the State Key Laboratory of Information Engineering in Surveying Mapping and Remote Sensing, Wuhan University, Wuhan 430079, China (e-mail: jias@whu.edu.cn; xuzq@whu.edu.cn).}
\thanks{X. Li and C. Hu are with the Lane Department of Computer Science
and Electrical Engineering, West Virginia University, Morgantown, WV 26506
USA (e-mail: xin.li@mail.wvu.edu, chuanbo.hu@mail.wvu.edu)}
\thanks{Corresponding author: Xin Li.}}
\markboth{IEEE Transactions on Information Forensics and Security}%
{Shell \MakeLowercase{\textit{et al.}}: Bare Demo of IEEEtran.cls for IEEE Journals}
%

\maketitle

\begin{abstract}
We have witnessed rapid advances in both face presentation attack models and presentation attack detection (PAD) in recent years. Compared to widely studied 2D face presentation attacks (e.g. printed photos and video replays), 3D face presentation attacks are more challenging because face recognition systems (FRS) is more easily confused by the 3D characteristics of materials similar to real faces. Existing 3D face spoofing databases, mostly based on 3D facial masks, are restricted to small data size and suffer from poor authenticity due to the difficulty and expense of mask production. In this work, we introduce a wax figure face database (WFFD) as a novel and super-realistic 3D face presentation attack. This database contains 2300 image pairs (totally 4600) and 745 subjects including both real and wax figure faces with high diversity from online collections. On one hand, our experiments have demonstrated the spoofing potential of WFFD on three popular FRSs. On the other hand, we have developed a multi-feature voting scheme for wax figure face detection (anti-spoofing), which combines three discriminative features at the decision level.  The proposed detection method was compared against several face PAD approaches and found to outperform other competing methods. Surprisingly, our fusion-based detection method achieves an Average Classification Error Rate (ACER) of 11.73\% on the WFFD database, which is even better than human-based detection.
\end{abstract}


\IEEEpeerreviewmaketitle

\section{Introduction}
\IEEEPARstart{F}{ace} has been one of the most widely-used biometrics modalities due to its accuracy and convenience for personal verification and identification. However, the increasing popularity and easy accessibility of face modalities also makes face recognition systems (FRS) a major target of spoofing such as presentation attack~\cite{ISO2017}. This kind of security threats can be easily performed by presenting the FRS a face artifact, which is also known as presentation attack instrument (PAI) in the ISO standard~\cite{ISO2016}. A recent breach of biometrics database (BioStar) leads to the compromise of as many as 28 million record containing facial recognition and fingerprint data, which can be easily exploited as PAIs by malicious hackers.
 
\begin{figure}[t]
\begin{center}
\subfigure[]{\label{(a)}
\includegraphics[width=1.21in,height=0.962in]{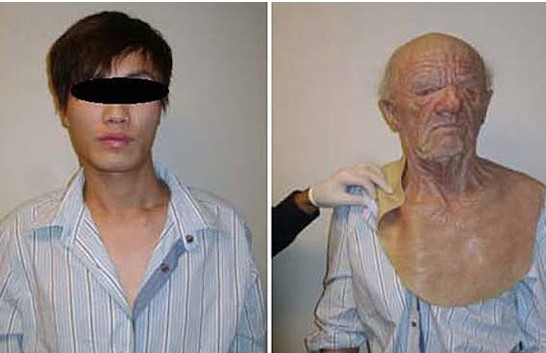}}
\subfigure[]{\label{(b)}
\includegraphics[width=1.0in,height=0.962in]{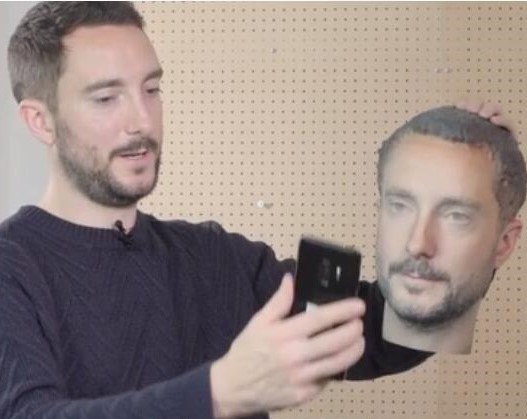}}
\subfigure[]{\label{(c)}
  \includegraphics[width=1.03in,height=0.962in]{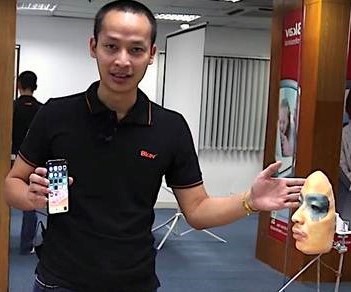}}
\caption{Examples of 3D presentation attack cases. (a) Airport security system fooled by silicon mask \protect\footnotemark[1], (b) Android phones fooled by a 3D-printed head\protect\footnotemark[2], (c) iPhone X face ID unlocked by a 3D mask\protect\footnotemark[3].}
\label{fig:1}  
\end{center}
\end{figure}

Based on the way of generating face artifacts, face presentation attacks can be classified into 2D modalities (e.g., printed/digital photographs or recorded videos on mobile devices such as a tablet) and 3D type (e.g., by wearing a mask or presenting a synthetic model). Existing systems and research on face recognition pay more attention to 2D face PAI due to its simplicity, efficiency, and low cost. However, as material science and 3D printing advance, creating face like 3D structures or materials has become easier and more affordable. When compared against the 2D modalities, 3D face presentation attacks are more realistic and therefore more difficult to be detected. The class of 3D face presentation attacks includes wearing wearable facial masks~\cite{3DMAD2013spoofing}, building 3D facial models~\cite{xu2016virtual}, through make-up~\cite{chen2017spoofing}, and using plastic surgery. Fig. \ref{fig:1} shows several examples of 3D presentation attacks, which have successfully fooled some widely used FRS such as in airport and phones. 

Existing research on 3D face presentation attacks focuses more on easy-to-make facial masks. 3D facial mask spoofing had been previously thought of impossible to become a common practice in the literature~\cite{zhang2012face} because 3D masks were deemed much more difficult and expensive to manufacture (e.g., requiring special 3D devices and materials). However, rapid advances in 3D printing technologies and services have made it easier and cheaper to make 3D masks recently. Several 3D mask attack databases have already been created, including 3D Mask Attack Database (3DMAD)~\cite{3DMAD2013spoofing}, 3D-face spoofing database (3DFS-DB)~\cite{galbally2016three}, HKBU 3D Mask Attack with Real World Variations Database (HKBU-MARs)~\cite{liu20163d}, Silicone Mask Attack Database (SMAD)~\cite{manja2017detecting}, and Wide Multi Channel Presentation Attack database (WMCA)~\cite{george2019biometric}.

\footnotetext[1]{Picture is downloaded from https://chameleonassociates.com/security-breach/.}
\footnotetext[2]{Picture is downloaded from http://www.floridaforensicscience.com/broke-bunch-android-phones-3d-printed-head/.}
\footnotetext[3]{Picture is downloaded from https://boingboing.net/2010/11/05/young-asian-refugee.html.}

These 3D face presentation attack databases have collected different 3D masks from the third-party services~\cite{3DMAD2013spoofing, liu20163d}, self-manufacturing~\cite{galbally2016three}, or online resources~\cite{manja2017detecting}. However, the databases are restricted to small data sizes (mostly less than 30 subjects), low mask qualities (some are not user customized~\cite{manja2017detecting, agarwal2017face}), low diversity in lighting conditions, facial poses and recording devices. These restrictions will greatly limit not only the attack abilities of fake faces but also the validity of research findings about detection performance against 3D presentation attacks. 

\begin{figure}[t]
\begin{center}
\subfigure[]{\label{(a)}
\includegraphics[width=1.33in,height=1.062in]{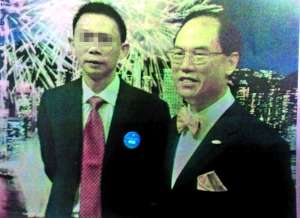}}
\subfigure[]{\label{(b)}
 \includegraphics[width=1.33in,height=1.062in]{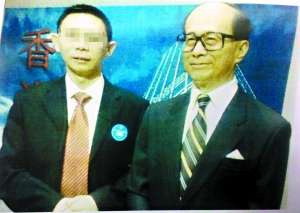}}
\caption{Photos with wax figure faces for fraud\protect\footnotemark[4]. (a) with the wax figure of Hong Kong chief executive Donald Tsang Yam-Kuen, (b) with the wax figure of Hong Kong business tycoon Li Ka-Shing.}
\label{fig:2}  
\end{center}
\end{figure}

To address these limitations, we propose to take advantage of the popularity and publicity of numerous celebrity wax figure museums in the world, and collect a large number of wax figure images to create a new Wax Figure Face Database (WFFD). These life-size wax figure faces are all carefully designed and made in clay with wax layers, silicone or resin materials, so that they are super-realistic and similar to real faces. With the development of wax figure manufacture technologies and services, we believe easily obtainable and super-realistic wax figure faces will pose threat to the existing face recognition systems. In fact, the wax figure faces have already been used for identity personation and fraud in real life. In 2012, using photos taken with the wax figures at Hong Kong’s Madam Tussauds Museum (as shown in Fig. \ref{fig:2}), six suspects snapped about 600,000 people out of nearly US\$475 million under a pyramid sales scam by claiming that their company was supported by Hong Kong chief executives and business tycoons.

\footnotetext[4]{Picture is downloaded from http://www.szdaily.com/content/2012-03/26/content\_6594713.htm.}

In this paper, we introduce these wax figure faces as a more challenging type of 3D face presentation attack and analyze their impact on face recognition systems. The main contributions of this work are summarized below. 
\begin{itemize}
\item The new WFFD database is constructed. It consists of 2300 images acquired from 745 subjects (with both real and wax figure faces, totally 4600 faces), which are diversified in terms of age, ethnicity, pose, expression, environment, and cameras. To the best of our knowledge, this is the first large-scale wax figure face database and it has not been proposed as super-realistic 3D face presentation attacks in the open literature.
\item Three classes of  discriminative features, including those learned from SqueezeNet and ResNet-50, and multi-block LPQ (Local Phase Quantization) texture feature, are extracted for wax figure face presentation attack detection. In view of their complementary nature, different feature fusion schemes are explored and compared. In particular, a novel multi-feature voting framework based on decision level fusion is proposed and its effectiveness is verified. 
\item We have conducted extensive experiments on the WFFD to justify its strong attack (spoofing) ability on three popular face recognition systems and several face PAD methods. The effectiveness of the proposed wax figure face detection (anti-spoofing) method is also demonstrated with comparison to previous state-of-the-art and human-based detection methods.
\end{itemize}

The rest of this paper is organized as follows. In Section II, we briefly review related research in 3D face presentation attack databases and PAD methods. The new WFFD database and three newly designed protocols are introduced in Section III. Section IV presents the proposed multi-feature voting detection scheme, and experimental results are reported in Section V. Finally, we make several conclusions about this work and future research in Section VI.
\section{Related work}

\subsection{Spoofing: 3D face presentation attack databases}
Existing 3D face presentation attack databases create attacks mainly based on wearable 3D face masks that are made of materials with face characteristics similar to real faces. 3DMAD~\cite{3DMAD2013spoofing} is the first publicly available 3D mask database. It used the services of ThatsMyFace\footnote[5]{http://thatsmyface.com/.} to manufacture 17 masks of users, and recorded 255 video sequences with an RGB-D camera of Micsoft Kinect device for both real access and presentation attacks. This database has been widely used since it provides not only color images and depth images but also manually annotated eye positions for all face samples. 

With the development of 3D modeling and printing technologies, more mask databases have been created since 2016. 3DFS-DB~\cite{galbally2016three} is a self-manufactured and gender-balanced 3D face spoofing database, in which 26 printed models were made using two 3D printers: the ShareBot Pro and the CubeX\footnote[6]{https://www.sharebot.it. and  http://www.cubify.com.}, which are relatively low-cost and worth about 1,000 and 2,000\texteuro, respectively. 
HKBU-MARs~\cite{liu20163d} is another 3D mask spoofing database with more variations to simulate real world scenarios. It generated 12 masks from two companies (ThatsMyFace and REAL-F\footnote[7]{http://real-f.jp/en\_the-realface.html.}) with different appearance qualities. A total of 1008 videos were created with 7 camera types and 6 lighting settings. To include more subjects,  SMAD database~\cite{manja2017detecting} has collected and compiled videos of people wearing silicone masks from online resources. It contains 65 genuine access videos of people auditioning, interviewing, or hosting shows, and 65 attacked videos of people wearing a complete 3D (but not customized) mask which fits well with proper holes for the eyes and mouth. 

Besides, there have been some 3D mask spoofing databases with special lighting information for more effective detection. The BRSU Skin/Face/Spoof Database~\cite{steiner2016reliable} provides multispectral SWIR (Short Wave Infrared) of four wavebands and RGB color images incorporating various types of masks and facial disguises. It contains 137 subjects and considers two face presentation attack scenarios: disguise of the own identity and counterfeiting of a foreign identity with a mask made of silicon, plastic, latex or hard resin materials. 

The MLFP database~\cite{agarwal2017face} (Multispectral Latex Mask based Video Face Presentation Attack database) is another multispectral database for face presentation attacks using latex and paper masks. It contains 1350 videos of 10 subjects in visible, near infrared (NIR), and thermal spectrums, which are captured at different locations (indoor and outdoor) in an unconstrained environment. The ERPA database~\cite{bhat2017you} provides the RGB and NIR images of both bona fide and 3D mask attack presentations captured using special cameras. This is a small dataset with only 5 subjects involved; the depth information is also available. Both rigid resin-coated masks and flexible silicone masks are considered. Similarly, the recently released WMCA database~\cite{george2019biometric} also used multiple capturing devices/channels, including color, depth, thermal and infrared. It contains 1679 videos with 347 bonafide and 1332 attacks from 72 subjects. A variety of 2D and 3D presentation attacks are included. For the 3D face attacks, it used fake head, rigid mask, flexible silicone masks, and paper masks and produced totally 709 videos.

These databases have played a significant role in designing multiple detection schemes against 3D face presentation attacks. However, they face the problems of small database size (mostly less than 30 subjects), poor authenticity (some based on paper or using noncustomized masks~\cite{manja2017detecting, agarwal2017face}), or low diversity in subject and recording process, which will certainly limit the development of effective and practical PAD schemes. 

\subsection{Anti-spoofing: 3D face PAD methods}
Detection of 3D fake faces is often more challenging than detecting fake faces with 2D planar surfaces. Existing PAD methods for 3D face presentation attacks are mainly based on the difference between real face skin and mask materials, which can be broadly classified into five categories- namely, reflectance based, texture based, shape based, liveness based, and deep features based.

Earlier studies~\cite{kim2009masked, zhang2011face, wang2013new, steiner2016reliable} in 3D mask spoofing detection were based on the reflectance difference of facial skins and mask materials. For example, the distribution of albedo values for illumination of various wavelengths was first analyzed in~\cite{kim2009masked} to find how different facial skins and mask materials (silicon, latex, and skinjell) behave in terms of reflectance. Then a 2D feature vector consisting of two average radiance values under 850nm (to distinguish between skins and mask materials) and 685nm (to distinguish different facial skin colors) was constructed. Using Fisher’s linear discriminant (FLD) classifier, this method ~\cite{kim2009masked} achieved 97.78\% accuracy in fake face detection on their own experimental data. 

Texture based methods explore the texture pattern difference of real faces and masks with the help of texture feature descriptors, such as the widely used Local Binary Patterns (LBP)~\cite{MORPHO1kose2013, MORPHO2kose2013shape,erdog2014spoofing, 3DMAD2013spoofing}, the Binarized Statistical Image Features (BSIF)~\cite{raghavendra2014novel, naveen2016face}, and Haralick features~\cite{agarwal2016face}. These methods are easy to implement, but their robustness to different mask spoofing attacks calls for further investigations. For example, different LBP features were tested in ~\cite{liu20163d} on their proposed database (HKBU-MARs), and it was found that LBP based methods can not generalize well when confronting different mask appearance.

Shape based 3D mask PAD methods use shape descriptors~\cite{MORPHO3kose2013vulnerability, tang20173d, hamdan2017detection} or 3D reconstruction~\cite{wang2018face} to extract discriminative features from faces and 3D masks. Different from reflectance-based or texture-based detection methods, these schemes only requires standard color images without the need of special sensors. However, their detection performances rely on the 3D mask attack qualities, and may not be roust to super-realistic 3D face presentation attacks.

More recently, some methods explore liveness cues to detect 3D face presentation attacks, such as thermal signatures~\cite{bhat2017you}, gaze information~\cite{ali2017biometric, alsufyani2018biometrie, ali2018gaze}, and pulse or heartbeat signals~\cite{liu20163d2, liu2018remote, hern2018time, Li2017Generalized}. Based on the intrinsic liveness signals, these methods achieve an outstanding performance on distinguishing real faces from masks.

Instead of extracting hand-crafted features, deep feature based methods automatically extract features from face images. Two deep representation approaches were investigated in~\cite{menotti2015deep} for detecting spoofing in different biometric modalities. Image quality cues (Shearlet) and motion cues (dense optical flow) were fused in ~\cite{feng2016integration} using a hierarchical neural network for mask spoofing detection, which achieved a half total error rate(HTER) of 0\% on the 3DMAD database. A network based on transfer learning using a pre-trained VGG-16 model architecture is presented in ~\cite{lucena2017transfer} to recognize photo, video and 3D mask attacks. Based on the observation with the importance of dynamic facial texture information, a deep convolutional neural network based approach was developed in ~\cite{shao2017deep}. Both intra-dataset and cross-dataset evaluation on 3DMAD and their supplementary dataset indicated the efficiency and robustness of the proposed method. Overall, deep learning based methods are not only efficient in spoof detection but also capable of recognizing different face presentation attacks. 

\section{The wax figure face database}
To address the weaknesses in existing 3D face presentation attack databases, we introduce a novel super-realistic Wax Figure Face Database (WFFD) with a large size and diversity in this paper. We will elaborate on the data collection process and the design of evaluation protocols related to WFFD in this section.

\subsection{Data collection}
\begin{figure}[t]
\begin{center}
\subfigure[]{\label{(a)}
\includegraphics[height=3.9cm]{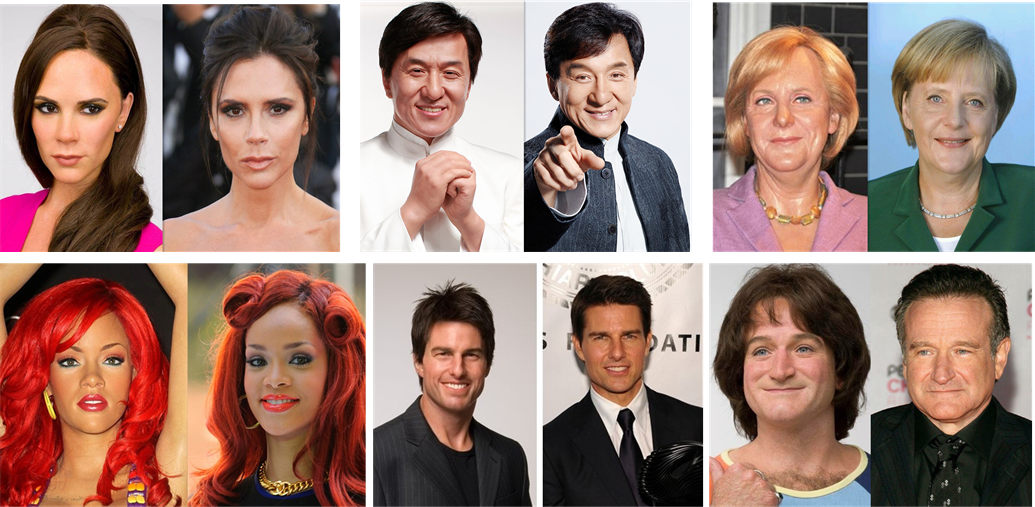}}
\subfigure[]{\label{(b)}
 \includegraphics[height=5.4cm]{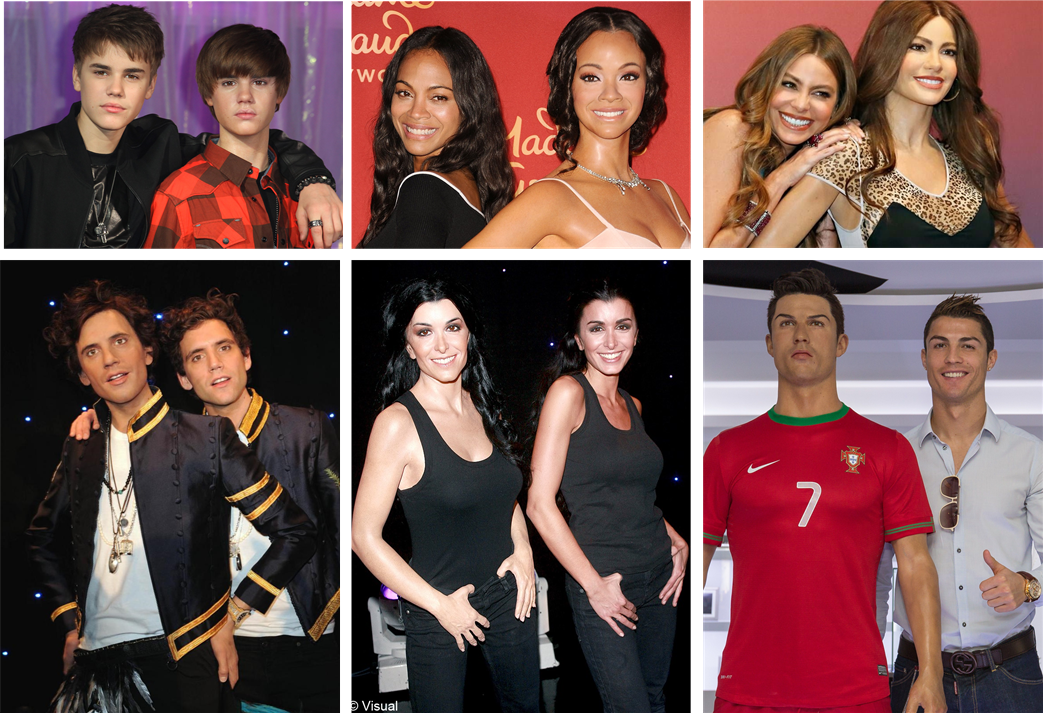}}
\caption{Image examples in the WFFD database. (a) Grouped manually, (b) recorded in the same scenario.}
\label{fig:3}  
\end{center}
\end{figure}

The images collected in the WFFD are based on numerous celebrity wax figure images from online resources. These user-customized and life-size wax figure faces are carefully designed and made in clay with wax layers, silicone or resin materials, so that they are super-realistic. We first downloaded as many celebrity wax figure faces as possible, and then collected the corresponding celebrity images as real access attempts. For each subject, the wax figure face and real face were finally grouped in one image to make a clear comparison, as the examples shown in Fig. \ref{fig:3}(a). In total, 1000 images were collected from 462 subjects.

Furthermore, we emphasize one particularly challenging scenario where the wax figure face and real person face were recorded together. Such scenario is only possible when the celebrities attended the unveiling of their own wax figures, as shown in Fig. \ref{fig:3}(b). Nevertheless, we have collected a total of 1300 images from 409 subjects for this challenging scenario. With the same recording environment and identical facial poses and expressions, these images are more difficult to distinguish even for humans. 

\subsection{Protocol design}
Overall, WFFD consists of 2300 images and 4600 faces from both real and wax figure faces of 745 subjects. Inspired by the fraud incident shown in Fig. \ref{fig:2}, we have further designed three situational protocols to evaluate the performance of face PAD methods on this database. 

1) Protocol I: \textit{heterogeneous}. This protocol contains images which are grouped manually. Since the wax figure face and real face are recorded from different devices and environmental conditions (e.g., lighting conditions), humans can make use of such subtle difference to distinguish the wax figure from the real face. 

2) Protocol II: \textit{homogeneous}. Images in this protocol record the wax figure face and real face in the same environment with the same camera. This situation is often challenging even for humans to tell wax figure apart from the real person.

3) Protocol III: \textit{mixed test}. The previous two protocols are combined to simulate real-world operational conditions. Note that with rapid advance in AI technology, it is possible to change the differing background in protocol I to make them appear ``homogeneous'' by image matting \cite{levin2007closed}.

In each protocol, images are grouped into three non-overlapped subsets: training, validation and testing. More details about the statistics of images in each protocol are shown in Table \ref{tab:1}.

\begin{table}[h]
\label{tab:1}
\newcommand{\tabincell}[2]{\begin{tabular}{@{}#1@{}}#2\end{tabular}} 
\small
\renewcommand{\arraystretch}{1.3}
\setlength{\tabcolsep}{4pt}
\caption{Details of each protocol in the WFFD}
  \centering
    \begin{tabular}{lcccccc}
\hline        
    \multirow{2}[3]{*}{\textbf{Protocol}} & \multicolumn{4}{c}{\textbf{\#Image}} & \multirow{2}[3]{*}{\textbf{\#Face}}&  \multirow{2}[3]{*}{\textbf{\#Subject}}\\
\cline{2-5}          & \textbf{Train} & \textbf{Valid} & \textbf{Test} & \textbf{Total} \\
     \hline
    Protocol I & 600   & 200   & 200   & 1000  & 2000  & 462 \\
     \hline
    Protocol II & 780   & 260   & 260   & 1300  & 2600  & 409 \\
     \hline
    Protocol III & 1380  & 460   & 460   & 2300  & 4600  & 745 \\
     \hline
    \end{tabular}%
\begin{tablenotes}
\scriptsize
\item[] Note that the train, validation, and test subsets are non-overlapped.
\end{tablenotes}
\end{table}

\subsection{Statistics} 
\begin{figure}[h]
\begin{center}
\includegraphics[width=3.45in]{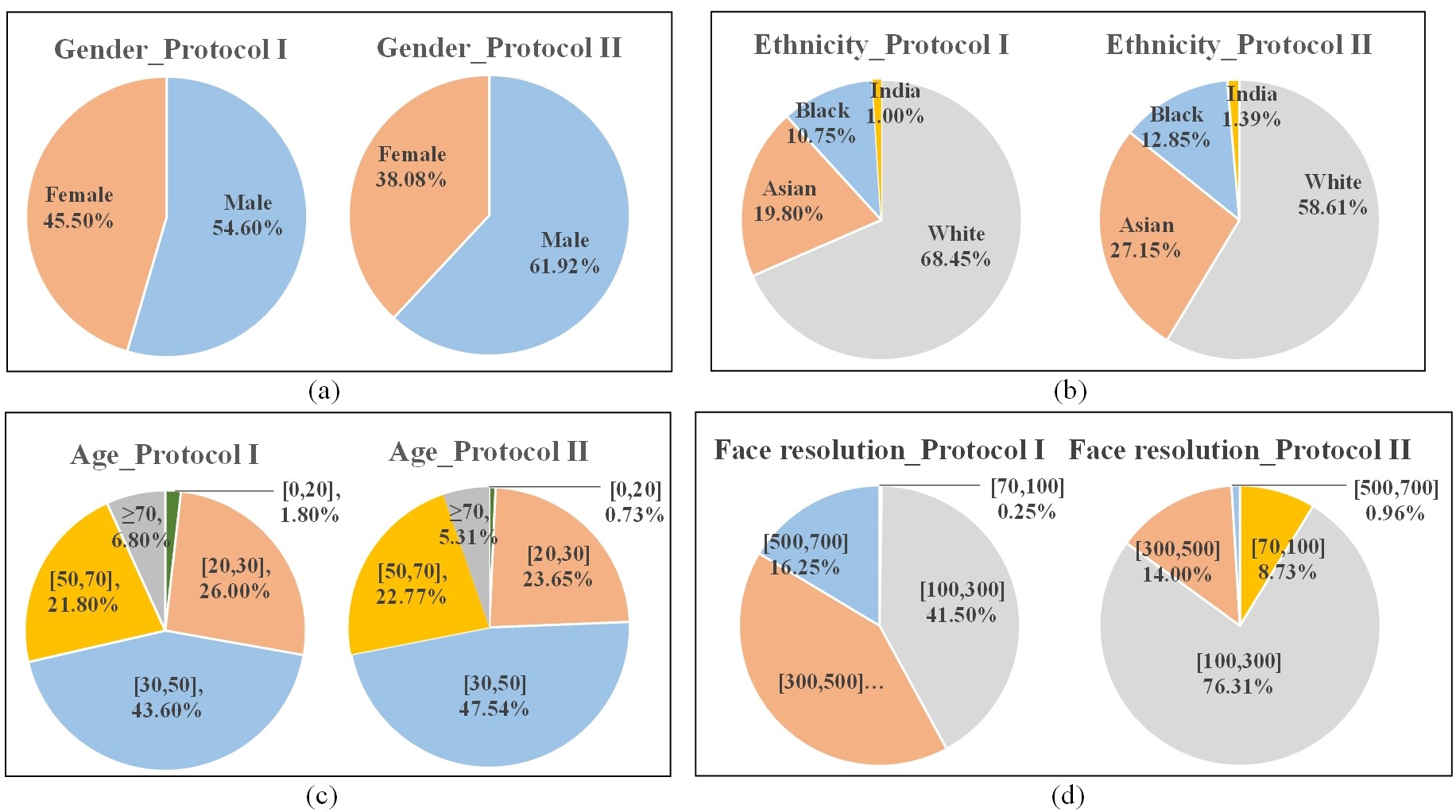}
\caption{Statistical distribution of the WFFD. (a) Gender, (b) ethnicity, (c) age, (d) face resolution.}
\label{fig:4}  
\end{center}
\end{figure}

The statistics information of  subject gender, age, ethnicity (detected by Face++~\cite{Facepp}), and face resolution (cropped by the dlib face detector~\cite{king2009dlib}) in the WFFD is shown in Fig. \ref{fig:4}. It can be seen that images in the WFFD are relatively gender balanced - with about 60\% of male and 40\% of female in both protocols. The ethnicity distribution in Fig. \ref{fig:4}(b) contains a majority of White subjects (around 60\%), followed by about 20\% Asians and 10\% Blacks, and a small percentage of Indians (no more than 2\%). We can also see a wide distribution of age in Fig. \ref{fig:4}(c). The two protocols have similar distribution patterns in terms of age, with half subjects being between 30 and 50 years old. 

Although the dimensions of most face regions are between $100\times100$ and $500\times500$, there is a big difference in the distribution between the two protocols. Matched and grouped manually, the dimensions of face regions in Protocol I are generally larger than those in Protocol II. Additionally, images in Protocol I are more diversified in terms of subject pose, facial expression, recording environment and devices than those in Protocol II. In summary, when compared with other 3D face presentation attack databases (as shown in Table II), our WFFD enjoys several advantages including large size, super reality and high diversity.

\begin{table*}[h]
\label{tab:2}
\begin{center}
\newcommand{\tabincell}[2]{\begin{tabular}{@{}#1@{}}#2\end{tabular}} 
\small
\renewcommand{\arraystretch}{1.3}
\setlength{\tabcolsep}{3pt}
\begin{threeparttable}   
\caption{Comparison of 3D face presentation attack datasets}
\begin{tabular}{lcccccc}
  \hline 
  \textbf{Database}&\textbf{Year}&\textbf{\#Subject}&\textbf{\#Sample}&\textbf{Format}&\textbf{Material}&\textbf{Description}\\   
  \hline 
  3DMAD~\cite{3DMAD2013spoofing} &2013&17&255 &video &paper, hard resin & 2D color images + 2.5D depth maps\\
   \hline
  3DFS-DB~\cite{galbally2016three}&2016&26 &520 &video &plastic & 2D, 2.5D images + 3D information\\ 
   \hline
  HKBU-MARs~\cite{liu20163d}&2016&12 &1008&video &/ &color images\\ 
     \hline
  BRSU~\cite{steiner2016reliable} & 2016 & 137 & 141 & image & {silicon, plastic,resin, latex} & multispectral SWIR, color images\\
   \hline  
  SMAD~\cite{manja2017detecting} &2017&/&130& video &silicone & color images, from online resources\\    
  \hline
  MLFP~\cite{agarwal2017face} &2017&10&1350 &video &latex, paper & visible, NIR, thermal images\\ 
   \hline
  ERPA~\cite{bhat2017you} &2017&5&86 &image &resin, silicone & RGB, thermal, NIR images + depth\\  
   \hline
  WMCA~\cite{george2019biometric} & 2019 & 72 & 1679 & video & rigid, silicone, paper& multiple channels of 2D, 3D attacks\\
  \hline
  WFFD (proposed)&2019 &745 & 2300 &image & wax figure & \tabincell{l}{color images, realistic, from online resources}\\
  \hline
\end{tabular}
\end{threeparttable}
\end{center}
\end{table*}

\section{Proposed method}
For the proposed color image based WFFD database, detection methods using reflectance properties, shape analysis, and liveness cues are simply not applicable. To distinguish wax figure faces from real faces, we have developed a multi-feature voting scheme based on deep learning models and texture descriptors. The overall scheme is decomposed of two steps: \emph{multi-feature detection} and \emph{decision-level voting}, which will be elaborated next.

\subsection{Mutli-feature detection}
Based on our previous work in~\cite{jia2019database}, we have found deep learning based methods are effective on discovering powerful feature representations for not only 2D face presentation attacks but also wax figure face detection. Based on this observation, two pre-trained deep neural networks are used to provide complementary and robust features for anti-spoofing purpose. One is based on the SqueezeNet~\cite{iandola2016squeezenet, boulkenafet2017compe}. It is mainly comprised of Fire modules, which consist of a squeeze convolution layer with only 1x1 filters, feeding into an expand layer that has a mix of 1x1 and 3x3 convolution filters (as shown in Fig. 5(a)). Fire modules and several pooling layers are then stacked to form a small network with reasonably high accuracy. The other is based on ResNet-50~\cite{he2016deep}. Deep Residual Networks (ResNets) are constructed by stacking residual units (see Fig. 5(b)). Thanks to the identity function introduced to the network, the gradient calculation in back-propagation can flow more effectively, which helps alleviate the notorious vanishing gradient problem \cite{bengio1994learning}. Indeed, ResNet-50 has excellent capability of discovering discriminative features and distinguishing fake faces from real ones~\cite{tu2017ultra, li2018exposing, liu2019multi}. 
 
\begin{figure}[h]
\begin{center}
\subfigure[]{\label{(a)}
\includegraphics[width=1.57in]{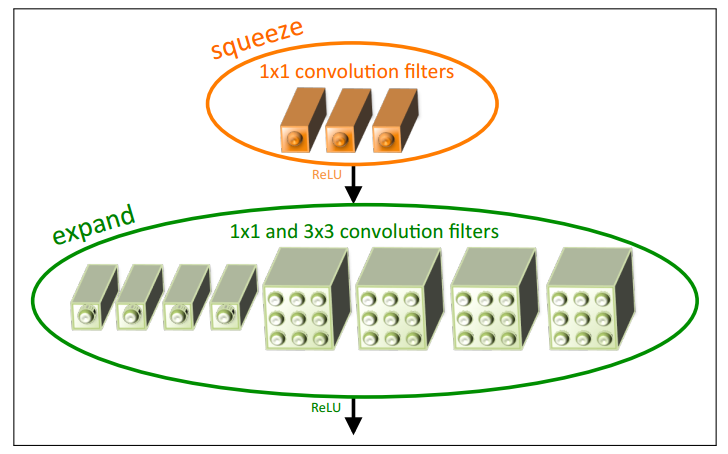}}
\subfigure[]{\label{(b)}
\includegraphics[width=1.65in]{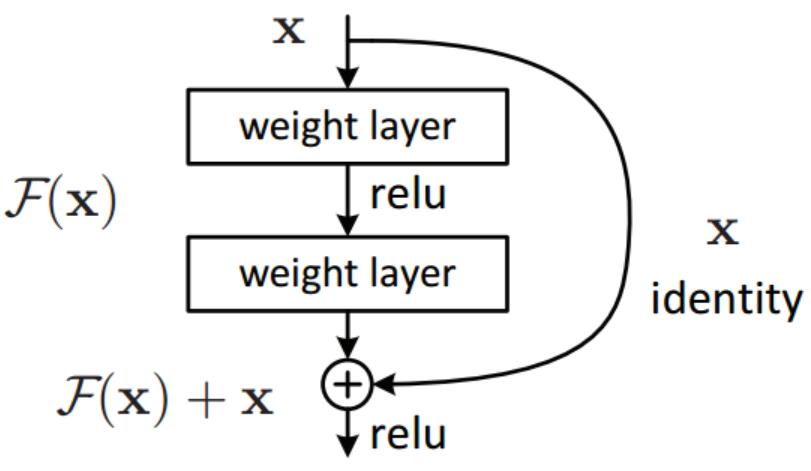}}
\caption{Building blocks in the two network architectures. (a) Fire module in SqueezeNet, (b) Residual learning in ResNet-50.}
\label{fig:5}  
\end{center}
\end{figure}

Due to the limited size of face presentation attack databases, it is often difficult to train a deep architecture from the scratch. Similar to  previous works~\cite{george2019biometric, boulkenafet2017compe, lucena2017transfer, tu2017ultra}, we propose to transfer the two deep neural networks (pre-trained using the celebrated ImageNet dataset) to fit the target database. They were fine-tuned on the training face images by the proposed WFFD database to avoid model overfitting. We first detect and resize all face images to $227\times227$ for SqueezeNet and $224\times224$ for ResNet-50 network as inputs. Formulating face PAD as a binary classification problem, we have removed the networks' output layer of 1,000 classes and obtained 1,000 dimensional output features of SqueezeNet (the feature dimension is 2,048 for ResNet-50). Then the features were fed into a Softmax classifier using cross-entropy loss function for optimization.

To further improve the detection accuracy, we have considered the inclusion of a traditional texture descriptor, namely multi-block local phase quantization (MB-LPQ), to characterize the intrinsic disparities in the color space of faces. Thanks to the discriminative power of local texture description, MB-LPQ has shown good performance in distinguishing real faces from artifacts in the literature ~\cite{jia2019database, boulkenafet2017compe, A11benl2015face}. In this work, we suggest that MB-LPQ is complementary to the two deep learning based features for wax figure face detection (which will be verified in the experiments in Section V). Taking the resized $64\times64$ face images as inputs, this feature detector converts standard RGB images into YCbCr color space and divides them into multiple blocks. The LPQ features extracted from each block are then concatenated to form the MB-LPQ feature vector, which is fed into a Softmax classifier for making the final prediction (real or fake).
\begin{figure*}[h]
\begin{center}
\includegraphics[width=6.05in]{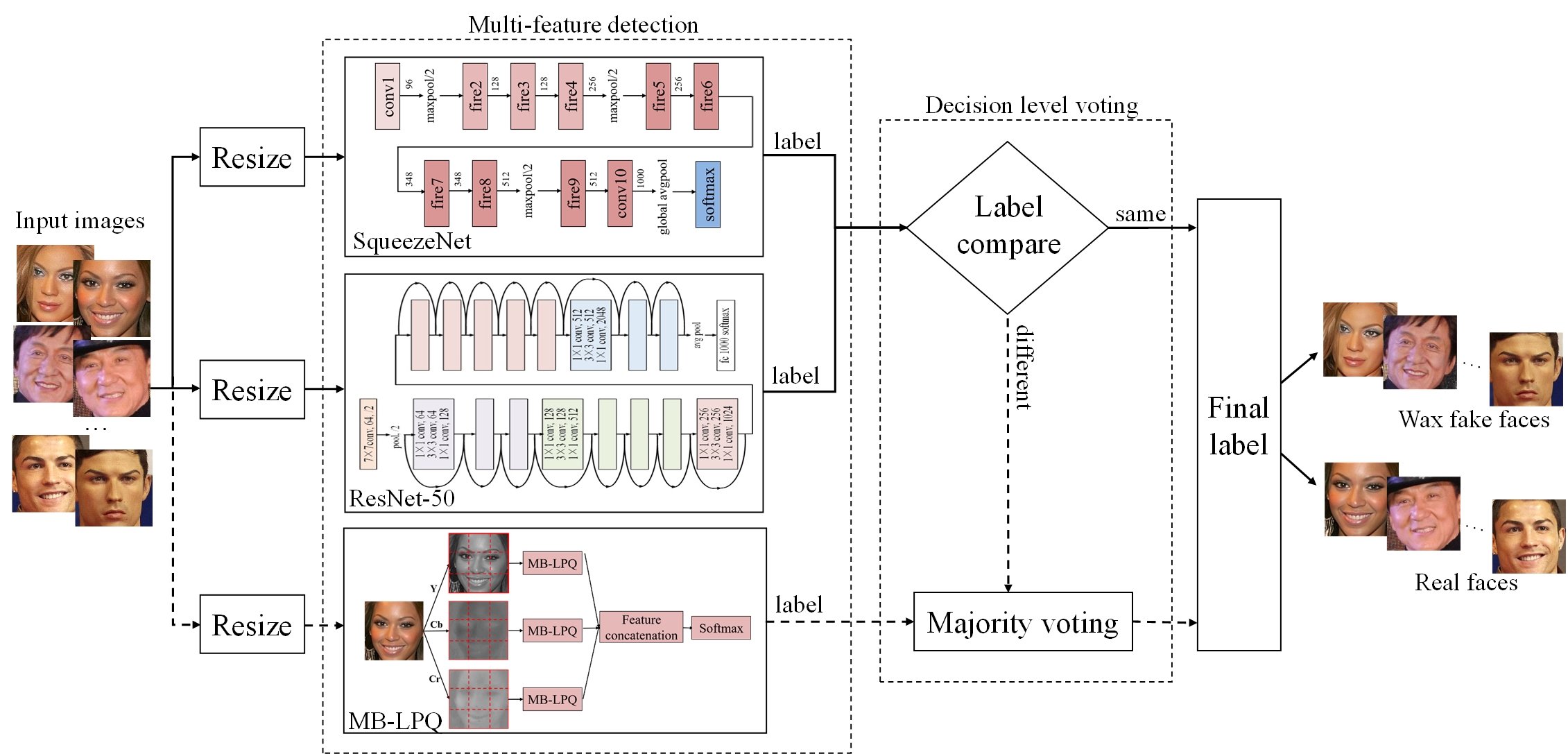}
\caption{Block diagram of the proposed detection scheme. Dashed lines indicate the optional process, which are performed only when `label compare' outputs `different'.}
\label{fig:6}  
\end{center}
\end{figure*}

\subsection{Decision-level voting}

The idea of combining classifiers dated back to \cite{kittler1998combining}. There are several ways of fusing the classification results - e.g., sum rule and majority voting. To exploit complementary features in our design, we propose a multi-feature voting scheme based on fusion at the decision level. Anti-spoofing detection labels predicted by two deep learning (SqueezeNet and ResNet-50) models are compared first. If they are the same, the consistent result will be directly output as the final predicted label; otherwise, different prediction results will be combined with the MB-LPQ feature detection result for further voting. The majority voting result of three competing models will be declared as the label of final prediction. The overall framework of the proposed multi-feature detection and combination scheme is shown in Fig. 6.

\section{Experimental Results}
In this section, we first demonstrate the attack ability of the introduced WFFD database by investigating the vulnerability of three popular face recognition systems to super-realistic 3D presentation attacks. Then the proposed multi-feature voting detection scheme is evaluated and compared with both human-based detection and several popular face PAD methods. Finally, we explore and analyze the failure cases of different anti-spoofing detection schemes.

\subsection{Evaluation metrics}
All experimental results were reported based on the ISO/IEC 30107-3 metrics~\cite{ISO2016}. For the evaluation of  vulnerability across different FRSs, the Impostor Attack Presentation Match Rate (IAPMR) metric was used. This metric has been widely used as an indicator of attack success probability if the FRS is evaluated against its PAD capabilities. It is defined by the proportion of impostor attack presentations using the same PAI species in which target reference is matched in a full-system evaluation of verification systems. For detection performance evaluation, we will calculate three types of errors - i.e.,  Attack Presentation Classification Error Rate (APCER),  Bona Fide Presentation Classification Error Rate (BPCER), and  Average Classification Error Rate (ACER). 

\subsection{Vulnerabilities of face recognition systems}
Three popular FRSs were considered in our experiments to demonstrate their vulnerability against fake/spoofing faces using the proposed WFFD database. To the best of our knowledge, this is the first study about the attack abilities of the super-realistic database on these popular FRSs. They are two publicly available FRSs: OpenFace~\cite{Openface} and Face++~\cite{Facepp}, and a commercial system Neurotechnology VeriLook SDK~\cite{Verilook}. Using the thresholds recommended by these FRSs, we have calculated the IAPMR values on three protocols of the WFFD, as presented in Table III.

\begin{table}[h]
\renewcommand{\arraystretch}{1.10}
\small
\renewcommand{\arraystretch}{1.3}
\setlength{\tabcolsep}{3pt}
\centering
\begin{threeparttable}   
  \caption{IAPMR of three Face Recognition Systems}
    \begin{tabular}{p{1.8cm}p{1.4cm}p{1.4cm}p{1.4cm}}
    \hline
    {Protocol}  & {Openface} &{Face++} &{VeriLook} \\
    \hline
    Threshold & 0.99\tnote{*} &1e-5\tnote{\dag} &36\tnote{**}\\
    \hline
    Protocol I& 93.20\%& 91.79\%  & 76.14\%  \\ 
    \hline
    Protocol II& 96.85\% & 96.22\% & 87.96\%  \\ 
    \hline
    Protocol III & 95.26\%  &94.30\% & 81.75\% \\
    \hline
\end{tabular}
\begin{tablenotes}
\scriptsize
\item[*] Using a squared L2 distance threshold; \dag Using the confidence threshold at the 0.001\% error rate; ** Using the matching score when FAR=0.1\%.
\end{tablenotes}
\end{threeparttable}
\end{table}

As shown in Table III, over 91\% of the images in the three protocols of the WFFD were successfully matched while using Openface or Face++, which implies the high attack success rates of the proposed WFFD. Meantime, lower values of  IAPMR can be observed for the VeriLook SDK. This is attributed to the fact that some wax figure faces with low qualities or special poses cannot be identified by the VeriLook SDK. However, we note that VeriLook tends to produce less successful matches than the other two even for real faces (refer to Fig. 7). 

In addition, by comparing the results between Protocol I and Protocol II, we can observe that higher matching rates were achieved for images in Protocol II (homogeneous). This is because FRSs are more easily fooled when fake faces and real faces are recorded by the same camera and even with identical facial expressions and poses. Such findings with IAPMR of $>96\%$ suggest that super-realistic WFFD could pose severe threat to existing FRSs without taking presentation attacks into account. 
\begin{figure*}[h]
\begin{center}
\subfigure[]{\label{(a)}
\includegraphics[width=2.27in]{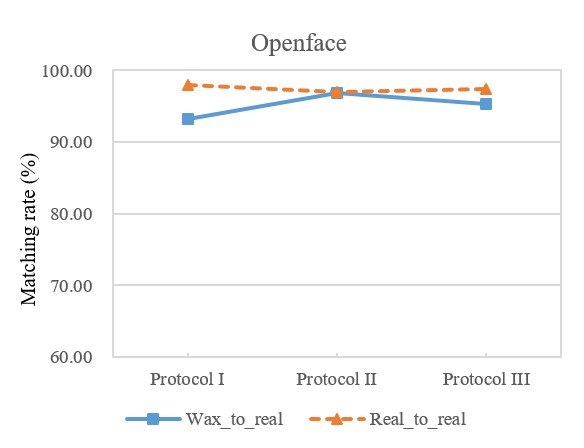}}
\subfigure[]{\label{(b)}
\includegraphics[width=2.27in]{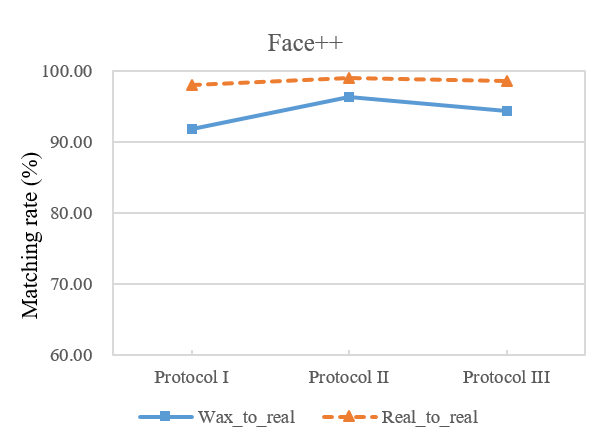}}
\subfigure[]{\label{(c)}
  \includegraphics[width=2.27in]{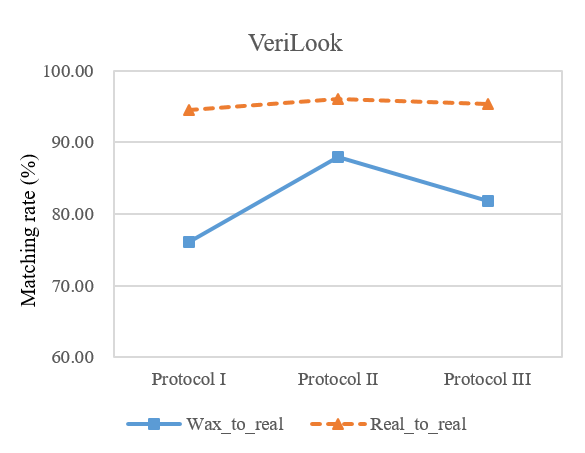}}
\caption{Comparison of matching rates in different face recognition systems. (a) Openface, (b) Face++, (c) VeriLook SDK.}
\label{fig:7}  
\end{center}
\end{figure*}
We have also compared the IAPMR values with the actual matching rates using real to real faces in Fig. 7. It can be seen that the matching rates using the wax figure faces to real faces are close to those using real to real faces for Openface and Face++ systems, which justifies the high fidelity of proposed WFFD. The gap between wax to real and real to real face matching is slightly larger for the VeriLook SDK system, which implies that anti-spoofing capability is an overlooked performance metric in the existing FRSs.

\subsection{Detection performance of the proposed PAD scheme}
Built upon three discriminative features, we want to demonstrate the effectiveness of fusion in detecting wax figure faces from real ones on the WFFD dataset. In our experiments, three different fusion schemes, including the feature-level fusion, score-level fusion, and decision-level fusion, were compared along with the proposed multi-feature voting method (refer to Fig. 6). The overall comparison results in terms of the  ACER are shown in Table IV.

It can be seen that without any fusion, the learned features from SqueezeNet model achieved the lowest ACER of 15.33\% for Protocol III, indicating the best discriminative power in detecting wax figure faces. Combining the two features learned from SqueezeNet and ResNet-50 models at the feature or score level (based on the sum rule) slightly improves the performance; further combining them with MB-LPQ features results in lower ACERs, reaching around 11.73\% for Protocol III. Overall, decision level fusion of the three features showed the best results. Specifically, when compared against direct fusion at the decision level, the proposed scheme can achieve the lowest ACERs for all three protocols also with lower computational cost due to the multi-voting strategy. Meanwhile, we can observe that for most fusion schemes, the error rates in the Protocol II were higher that those in the Protocol I, suggesting that detecting the wax figure faces from real faces recorded in the homogeneous case is the most challenging.   

\begin{table}[htbp]
\newcommand{\tabincell}[2]{\begin{tabular}{@{}#1@{}}#2\end{tabular}} 
\small
\renewcommand{\arraystretch}{1.3}
\setlength{\tabcolsep}{3pt}
  \centering
  \caption{Comparison results (ACER) of different fusion schemes}
    \begin{tabular}{lp{2.2cm}lll}
    \hline
          \multicolumn{1}{c}{Fusion method} & \multicolumn{1}{c}{Feature} & \multicolumn{1}{c}{\tabincell{l}{Protocol\\I}} & \multicolumn{1}{c}{\tabincell{l}{Protocol\\II}} & \multicolumn{1}{c}{\tabincell{l}{Protocol\\III}} \\
    \hline
    \multirow{3}[0]{*}{Single feature} & F1-SqueezeNet & 16.25 & 14.61 & 15.33 \\
\cline{2-5}          & F2-ResNet-50 & 19.75 & 20.57 & 19.02 \\
\cline{2-5}          & F3-MB-LPQ & 17.50  & 31.35 & 22.28 \\
    \hline
   \multirow{4}[0]{*}{Feature level} & F1 \& F2 & 15.00  & 16.35 & 14.67 \\
\cline{2-5}          & F1 \& F3 & 16.00 & 17.31 & 15.11 \\
\cline{2-5}          & F2 \& F3 & 13.50  & 19.42 & 15.54 \\
\cline{2-5}          & F1 \& F2 \& F3 & 14.75 & 15.38 & 13.70 \\
    \hline
    \multirow{4}[0]{*}{Score level} & F1 \& F2 & 16.25 & 15.38 & 15.00 \\
\cline{2-5}          & F1 \& F3 & 16.00  & 20.38 & 18.26 \\
\cline{2-5}          & F2 \& F3 & 16.75 & 22.50 & 19.67 \\
\cline{2-5}         & F1 \& F2 \& F3  & 12.75 & 15.96 & 12.83 \\
    \hline
   \multirow{2}[0]{*}{Decision level} &  F1 \& F2 \& F3  &11.28  &14.23  &12.00  \\
\cline{2-5}          & {\tabincell{l}{Proposed: F1 \&\\F2 (\&F3)}} & \textbf{11.25} & \textbf{13.65} & \textbf{11.73} \\
    \hline
    \end{tabular}%
  \label{tab:addlabel}%
\end{table}%

Fig. 8 shows the APCER and BPCER performance of different fusion schemes under three protocols of WFFD dataset. It can be observed from Fig. 8(a) that the detection performance (including both APCER and BPCER values) changed little for different fusion schemes under Protocol I. However, under the more challenging Protocol II in Fig. 8(b), all competing fusion schemes achieved higher BPCER than APCER. This suggests that more real faces were incorrectly classified as wax figure faces when they were homologous. Further, feature-level fusion leads to lower BPCER (around 20\%) while score-level fusion and decision-level fusion lead to lower APCER values (around 10\%). Similar trends can be observed in Fig. 8(c) under the comprehensive Protocol III. Overall, due to the high fidelity and large diversity of wax figure faces in WFFD, distinguishing wax faces from real faces is challenging especially for the homogeneous situation (Protocol II). By exploiting the the complement property among different features, the proposed multi-voting fusion scheme has achieved the lowest APCER of 7.8\% and the lowest BPCER of around 15\%.

\begin{figure*}[h]
\begin{center}
\subfigure[]{\label{(a)}
\includegraphics[width=2.3in,height=1.6in]{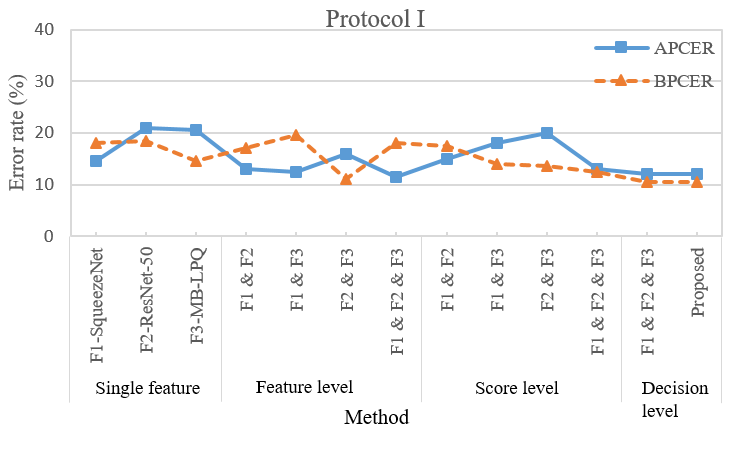}}
\subfigure[]{\label{(b)}
\includegraphics[width=2.3in,height=1.6in]{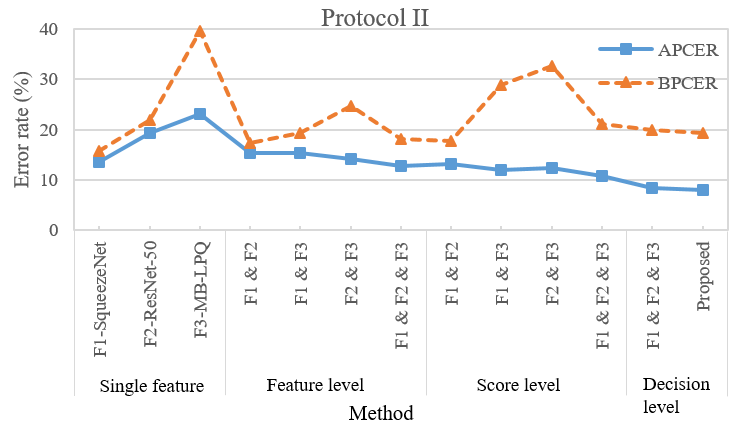}}
\subfigure[]{\label{(c)}
  \includegraphics[width=2.3in,height=1.6in]{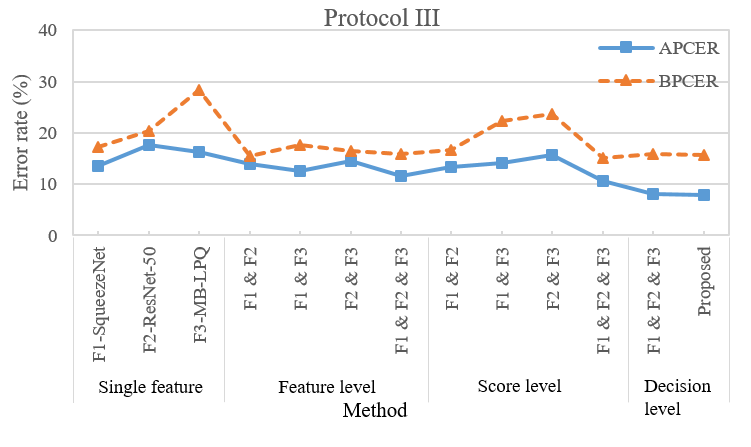}}
\caption{Comparison results of different fusion schemes under three protocols. (a) Protocol I, (b) Protocol II, (c) Protocol III.}
\label{fig:8}  
\end{center}
\end{figure*}

\subsection{Comparison against other PAD methods}
Several face PAD methods were evaluated and compared on the WFFD database to show how they can work for super-realistic 3D presentation attacks. These PAD methods have achieved promising performance in detecting 2D type or 3D mask presentation attacks based on different features. Our benchmark set includes multi-scale LBP~\cite{3DMAD2013spoofing}, the reflectance properties based~\cite{MORPHO4kose2013reflectance}, image quality assessment based~\cite{A7galba2014face}，color LBP~\cite{A12boul2015face},  Haralick features~\cite{agarwal2016face},  VGG-16 model based~\cite{lucena2017transfer}, Chromatic Co-Occurrence of LBP (CCoLBP)~\cite{peng2018ccolbp}, and noise modeling based~\cite{jourabloo2018face}. The experimental results of all benchmark methods were obtained using their public available codes. In addition, we have conducted a controlled human-based detection experiment to test the ability of human eyes in distinguishing wax figure faces from real ones. In our controlled experiment, 20 volunteers (10 men and 10 women, aged between 23 and 55) were asked to determine whether the face is real or not using our self-developed program (as shown in Fig. 9). The classification error rates were calculated and then their average is taken as the final human-based detection result. 

\begin{figure}[h]
\begin{center}
\includegraphics[width=1.85in]{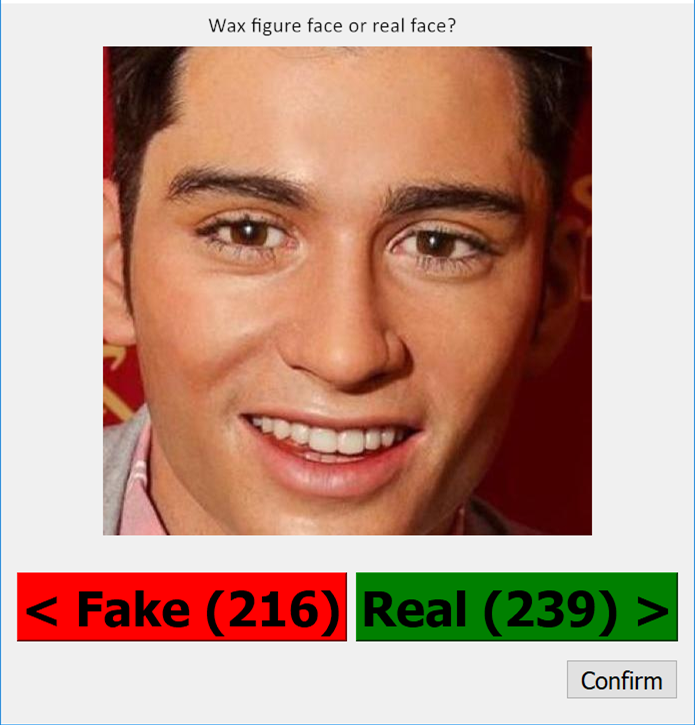}
\caption{The program for human-based detection.}
\label{fig:9}  
\end{center}
\end{figure}

Table V compares the results of different detection schemes. For Protocol I, we can see that existing face PAD methods for 2D or 3D mask attacks suffered from severe performance degradation with high detection error rates on WFFD, ranging from 26\% to 46\%. We attribute the poor performances to high diversity and super-realistic attacks in the introduced database. Human based detection has achieved better result with ACER of 16\%, but with higher APCER than BPCER, suggesting that more wax figure faces were mistaken for real ones. Our proposed multi-voting scheme achieved the best result with 11.25\% ACER. Similar performance differences can be observed under the Protocol II. However, most algorithms achieved higher error rates for this protocol. Such results are reasonable since recording the real faces and wax figure faces in the same scenarios with the same camera results in less difference between real and fake faces. Therefore, it is more difficult to detect the presentation attacks in this homogeneous setting, which is consistent with the results in Table III. 

\begin{table*}[t]
\small
\renewcommand{\arraystretch}{1.3}
\setlength{\tabcolsep}{3pt}
\centering
\begin{threeparttable}   
\caption{Detection error rates (\%) on three protocols of the WFFD}
    \begin{tabular}{l|llll|llll|llll}
      \hline
    {\multirow{2}[0]{*}{Method}} & \multicolumn{4}{c|}{Protocol I} & \multicolumn{4}{c|}{Protocol II}& \multicolumn{4}{c}{Protocol III} \\
      \cline{2-13} 
 & \multicolumn{1}{l}{EER} & \multicolumn{1}{l}{APCER} & \multicolumn{1}{l}{BPCER} & \multicolumn{1}{l|}{ACER} & \multicolumn{1}{l}{EER} & \multicolumn{1}{l}{APCER} & \multicolumn{1}{l}{BPCER} & \multicolumn{1}{l|}{ACER}& \multicolumn{1}{l}{EER} & \multicolumn{1}{l}{APCER} & \multicolumn{1}{l}{BPCER} & \multicolumn{1}{l}{ACER} \\
    \hline
    Multi-scale LBP~\cite{3DMAD2013spoofing}&33.17  & 31.22 &  31.22  & 31.22&36.62  & 37.32  & 33.45  & 35.39& 34.56 &  33.33  & 32.92  & 33.13\\ 
   \hline
   Reflectance~\cite{MORPHO4kose2013reflectance} &41.95  & 40.00  & 52.19 &  46.10 &44.37 &  50.70  & 44.37 &  47.53&  44.78&   46.01 &  46.22  & 46.11\\ 
   \hline
   Image quality~\cite{A7galba2014face} & 35.50 &  30.50  & 39.50 &  35.00& 38.85 &  39.23  &43.46  & 41.34&	41.30  & 36.96 &  43.26 &  40.11\\ 
   \hline
    Color LBP~\cite{A12boul2015face} &  33.17 & 30.24 & 36.10  &33.17 & 37.32  & 36.62  & 41.90  & 39.26 & 36.81  & 35.38  & 35.79 & 35.58\\ 
   \hline
    Haralick features~\cite{agarwal2016face} & 32.19 & 25.85 &  37.07 &  31.46&38.38 & 41.55 & 24.65 & 33.10 &  36.81  & 36.40  & 32.92 &  34.66  \\ 
   \hline
    VGG-16 based~\cite{lucena2017transfer}&45.85 &  50.73  & 41.95  & 46.34& 48.94 & 40.14 & 52.82 &  46.48&48.67 &  45.19 &  49.28  & 47.24\\
  \hline
    CCoLBP~\cite{peng2018ccolbp} &29.50   &26.50   &26.00  & 26.25 &28.08  & 24.62& 34.23 & 29.42 &28.04  & 26.52 &  29.13  & 27.83\\
    \hline
     Noise modeling~\cite{jourabloo2018face} &52.50 &  63.00  & 45.50  & 54.25  &58.85 &  58.46 &  59.61  & 59.04 & 56.09 &  58.69  & 49.98  & 54.33\\
     \hline
     Human-based &/&20.14& 11.86&  16.00 &/&32.97&  \textbf{17.97}&   25.47 & /&27.39& \textbf{15.31}& 21.35 \\
     \hline
     \textbf{The proposed} & \textbf{11.50} & \textbf{12.00 } & \textbf{10.50}  & \textbf{11.25}  & \textbf{12.00}&  \textbf{8.08} &  19.23  & \textbf{13.65} & \textbf{11.67}& \textbf{7.82} &  15.64 &  \textbf{11.73}\\
       \hline
\end{tabular}
\end{threeparttable}
\end{table*}

The overall results under the Protocol III of different face PAD methods have large differences, with the error rates ranging from 7.82\% to 48.67\%. The best ACER was achieved in the proposed multi-voting fusion scheme due to the highly discriminative and complementary features, which significantly outperformed other algorithms and human based detection. In terms of the BPCER, the human based detection obtained the lowest value of 15.31\%, slightly better than the proposed method. Besides, the CCoLBP features \cite{peng2018ccolbp} also achieved better results, with all error rates lower than 30\%. 

\subsection{Failure case analysis}
Based on the detection results on WFFD, we further show and analyze the failure cases to have a deep understanding of both detection methods and database.
 
\textbf{Features used in the proposed method.} In Fig. 10, we have shown the Venn diagram for the failure cases of the three single features used in our method (SqueezeNet, ResNet-50, and MB-LPQ). It can be seen that they achieved different failure cases in detecting the 920 faces in the testing subset of Protocol III, but only 33 were wrongly classified by all three features, implying the good complementary properties among them. We have also shown these 33 failure cases in Fig. 10, which visually illustrate the challenge with distinguishing between fake faces and real ones even for human eyes. Additionally, we note that more real faces were mistaken for fake ones (highlighted by the green dots in Fig. 10); by contrast, only around one third of failure cases mistaken wax faces by real ones (marked by the red dots in Fig. 10).

\begin{figure}[h]
\begin{center}
\includegraphics[width=3.5in]{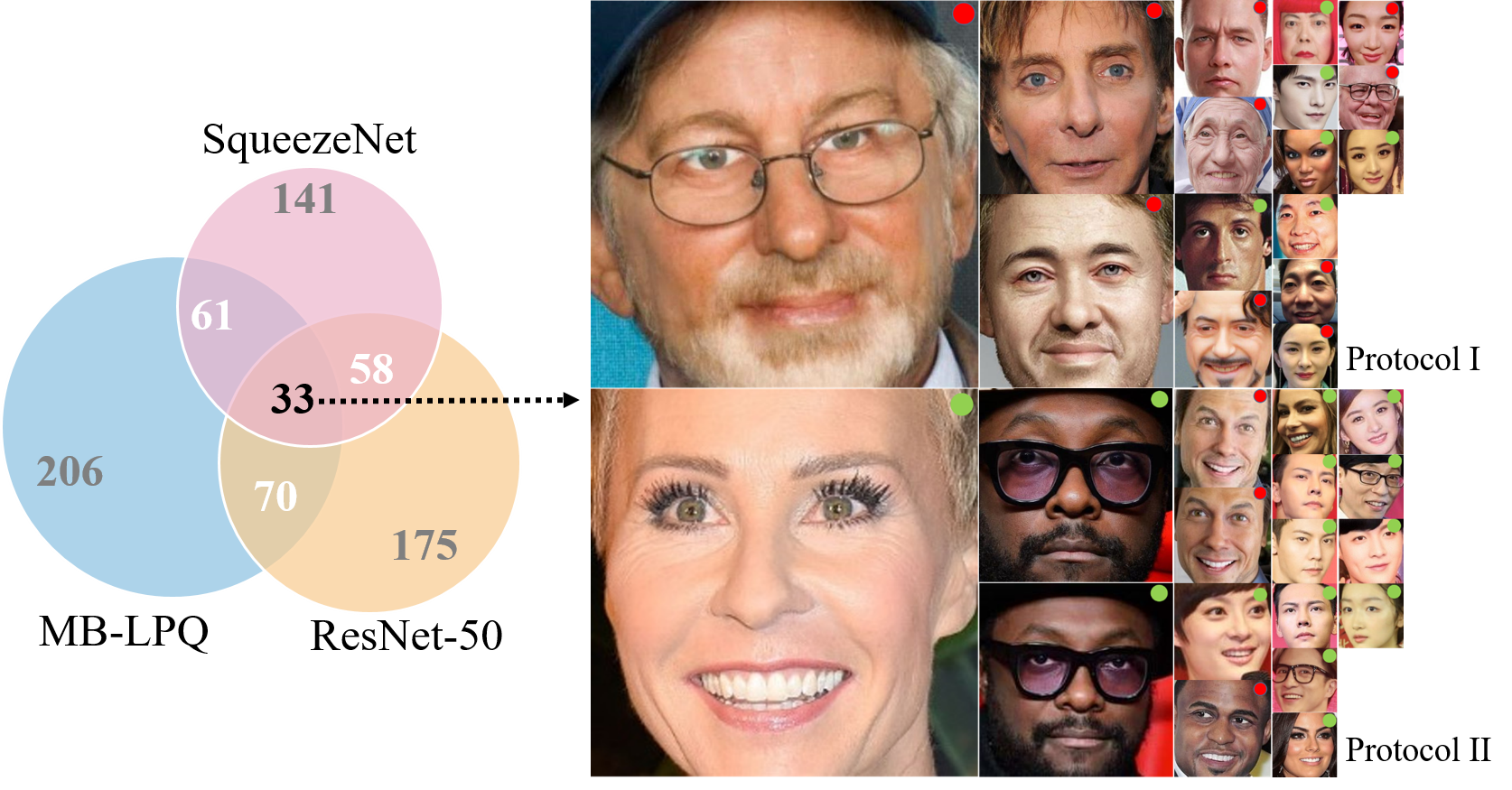}
\caption{The failure cases in feature based anti-spoofing detection. The left shows the Venn diagram of failure cases associated with three features; and the right shows the exemplar images in Protocols I and II. Note that images with green dots are real faces (but mistaken for wax faces), while images with red dots are wax figure faces (but mistaken for real faces).}
\label{fig:10}  
\end{center}
\end{figure}


\textbf{Human-based method.} We have also analyzed the detection results of 20 volunteers as shown in Table V. We make the following two observations. First, human-based detection performs worse than machine-based for all three protocols, which implies that real vs. wax detection is nontrivial for layperson. However, we note that this is partially due to the lack of training in 20 volunteers. With more experience, human observers tend to perform better in spoofing detection. Second and more interestingly, when compared against machine based detection, human based method was more likely to mistake wax figure faces for real ones for both protocols, as shown in Fig. 11 (there are more red dots than green dots). This is in sharp contrast with what we have observed for machine-based method in Fig. 10 (there are more green dots than red dots).
\begin{figure}[h]
\begin{center}
\includegraphics[width=3.3in]{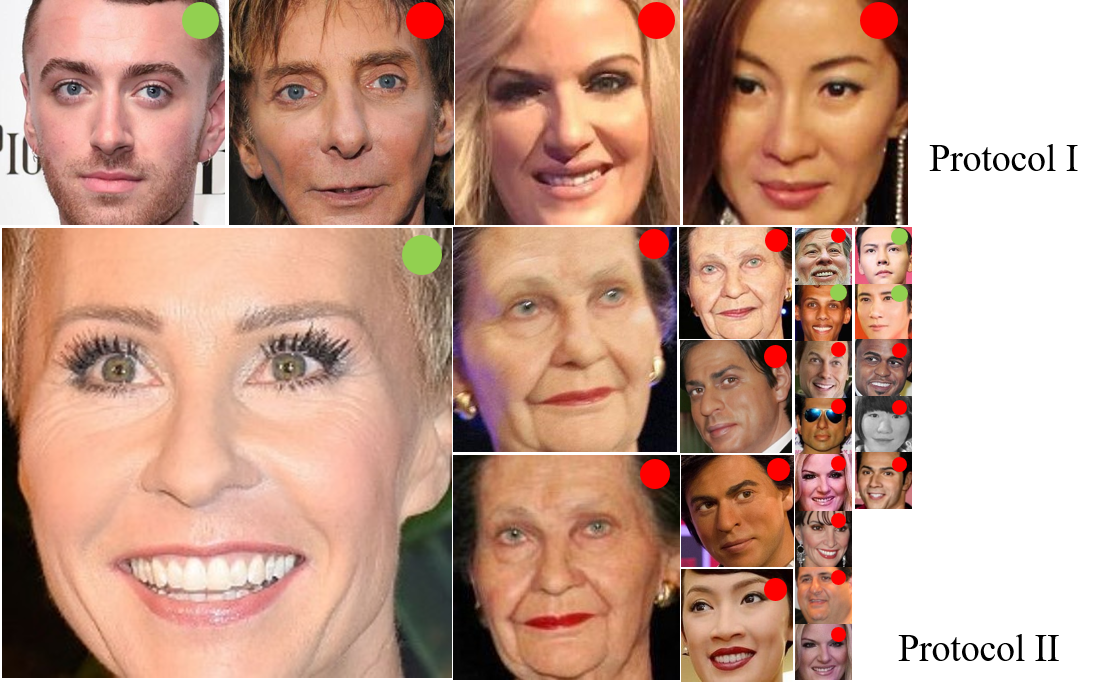}
\caption{The failure cases with high probabilities in human based anti-spoofing detection. Similarly, images with green dots are real faces (but mistaken for wax faces), while images with red dots are wax figure faces (but mistaken for real faces). Note that humans mistake more wax figure faces for real faces (red dots) than the other way around.}
\label{fig:11}  
\end{center}
\end{figure}

\section{Conclusions}
To address the limitations in existing 3D face presentation attack databases, we have constructed a new database (WFFD) composed of wax figure faces with high diversity and large size as super-realistic face presentation attacks. The database will be made publicly available to facilitate the improvement and evaluation of different PAD algorithms. Extensive experimental results have demonstrated the vulnerability of popular face recognition systems to these attacks. In particular, we have observed that several existing PAD methods fail in the task of detecting real faces from wax figure faces, demonstrating the challenges when wax figure face are used for 3D attacks. We have developed a multi-voting fusion scheme based on three discriminative and complementary features, which significantly outperformed not only current state-of-the art face PAD methods but also human-based detection. Through detailed analysis of failure cases, we have found that machine-based and human-based methods suffer from different types of errors. 

It should be noted that the best performance achieved by the proposed multi-voting scheme still has the error rate of over $10\%$. Super-realistic wax figure faces are indeed difficult to distinguish from real ones even for humans. We envision that motion based (instead of appearance based) anti-spoofing methods, such as head movement or blink detection, will deserve further study in the future. In view of recent advances in generative adversarial network (GAN)-based video synthesis (e.g., talking Mona Lisa and DeepFake), even motion based anti-spoofing might be foiled by more intelligent spoofing. And there have been already a flurry of works on detecting DeepFake \cite{agarwal2019protecting}, \cite{guera2018deepfake}, \cite{maras2019determining}. As many people believe, the arm races between spoof and anti-spoofing will never end. 

\section*{Acknowledgment}
Xin Li's work is partially supported by the DoJ/NIJ under grant NIJ 2018-75-CX-0032, NSF under grant OAC-1839909 and the WV Higher Education Policy Commission Grant (HEPC.dsr.18.5).

\ifCLASSOPTIONcaptionsoff
  \newpage
\fi

{\scriptsize
\bibliographystyle{IEEEtran}
\bibliography{egbib}
}





\end{document}